\documentclass[runningheads]{llncs}
\usepackage[T1]{fontenc}
\usepackage{multirow} 
\usepackage{cite}
\usepackage{hyperref}
\usepackage{amsfonts}

\usepackage{graphicx}
\begin{document}

\title{Why patient data cannot be easily forgotten?}
\titlerunning{Why patient data cannot be easily forgotten?}

\author{Ruolin Su\inst{1*}\and Xiao Liu\inst{1*} \and Sotirios A. Tsaftaris\inst{1,2}} 
\authorrunning{R. Su et al.}
\institute{School of Engineering, University of Edinburgh, Edinburgh EH9 3FB, UK 
\and
The Alan Turing Institute, London, UK\\
\email{R.SU-1@ed.ac.uk, Xiao.Liu@ed.ac.uk, S.Tsaftaris@ed.ac.uk}}

\maketitle          

\def\thefootnote{*}\footnotetext{Contributed equally}\def\thefootnote{\arabic{footnote}}
\setcounter{footnote}{0} 
\begin{abstract}
Rights provisioned within data protection regulations, permit patients to request that knowledge about their information be eliminated by data holders. With the advent of AI learned on data, one can imagine that such rights can extent to requests for forgetting knowledge of patient's data within AI models. However, forgetting patients' imaging data from AI models, is still an under-explored problem. In this paper, we study the influence of patient data on model performance and formulate two hypotheses for a patient's data: either they are common and similar to other patients or form edge cases, i.e.\ unique and rare cases. We show that it is not possible to easily \textit{forget patient data}. We propose a targeted forgetting approach to perform patient-wise forgetting. Extensive experiments on the benchmark Automated Cardiac Diagnosis Challenge dataset showcase the improved performance of the proposed targeted forgetting approach as opposed to a state-of-the-art method.

\end{abstract}

\section{Introduction}
Apart from solely improving algorithm performance, developing trusted deep learning algorithms that respect data privacy has now become of crucial importance \cite{abadi2016deep, liu2020have}. Deep models can memorise a user's sensitive information \cite{arpit2017closer, hartley2022unintended, jegorova2021survey}. Several attack types \cite{truex2019demystifying} including simple reverse engineering \cite{fredrikson2015model} can reveal private information of users. Particularly for healthcare, model inversion attacks can even recover a patient's medical images \cite{wu2020evaluation}. It is then without surprise why a patient may require that private information is not only deleted from databases but that any such information is forgotten by deep models trained on such databases.   

A naive solution to forget a patient's data is to re-train the model without them. However, re-training is extremely time-consuming and sometimes impossible \cite{shintre2019making}. For example, in a federated learning scheme \cite{mcmahan2017communication}, the data are not centrally aggregated but retained in servers (e.g.\ distributed in different hospitals) which may not be available anymore to participate in re-training.

\begin{figure}[t]
\centering
\includegraphics[width=0.95\textwidth]{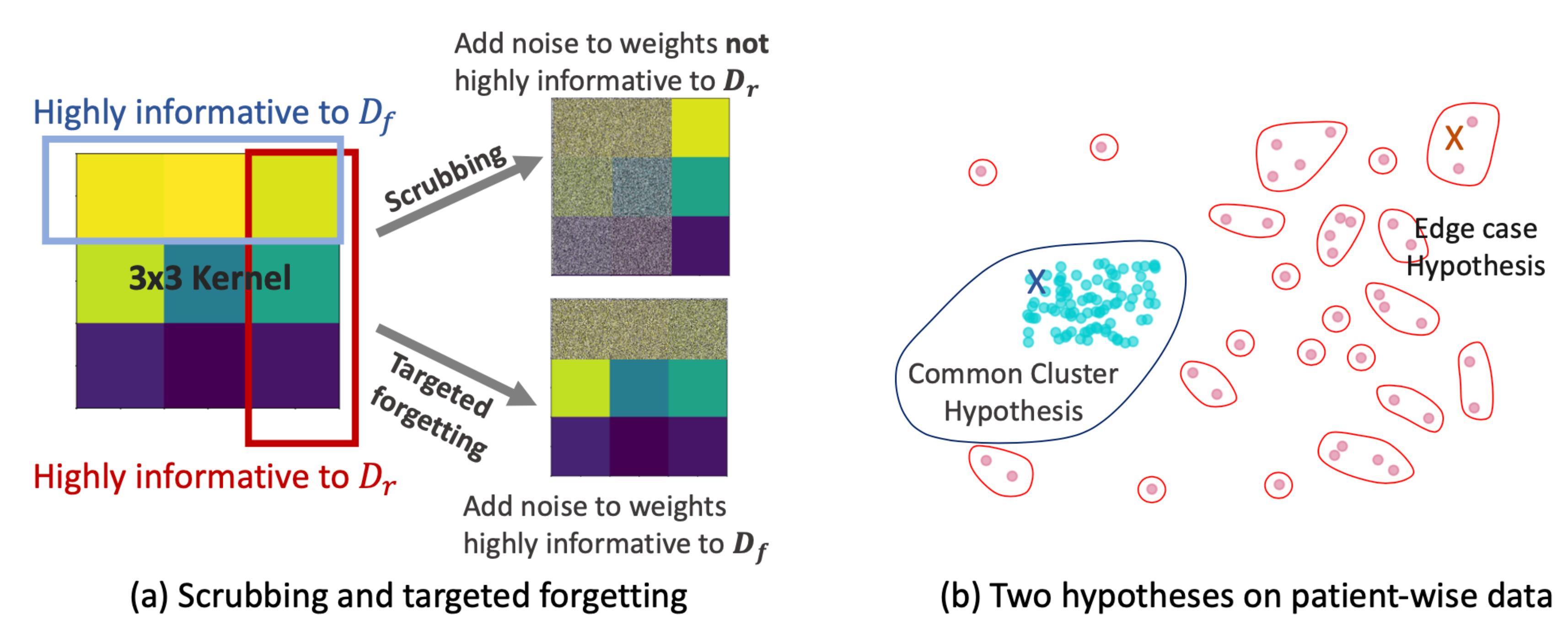}
\caption{(a) Visualisation of the scrubbing and targeted forgetting methods. $\mathcal{D}_{r}$ and $\mathcal{D}_{f}$ are the retaining data and the forgetting data. (b) Illustration of the two hypotheses. Blue contour delineates a big sub-population of similar samples within a \textit{common cluster}; red contours denote several small sub-populations of distinct samples in \textit{edge cases}. \color{blue}X\color{black}\ and \color{red}X\color{black}\ are examples of  samples to be forgotten.} 
\label{fig:1}
\end{figure}

As more advanced solutions, machine unlearning/forgetting approaches aim to remove private information of concerning data without re-training the model. This involves post-processing to the trained model to make it act like a re-trained one that has never seen the concerning data. Several studies have previously explored forgetting/unlearning data and made remarkable progress \cite{7163042,ginart2019making, golatkar2020eternal,nguyen2020variational,sekhari2021remember}. When the concept of machine unlearning/forgetting was first developed in \cite{7163042}, they discussed forgetting in statistical query learning \cite{kearns1998efficient}. Ginart et al. \cite{ginart2019making} specifically deal with data deletion in k-means clustering with excellent deleting efficiency. Another approach is to rely on variational inference and Bayesian models \cite{nguyen2020variational}. Recently, Sekhari et al. \cite{sekhari2021remember} propose a data deleting algorithm by expanding the forgetting limit whilst reserving the model's generalization ability. Golatkar et al. \cite{golatkar2020eternal} address machine unlearning on deep networks to forget a subset of training data with their proposed scrubbing procedure (shown in Fig.~\ref{fig:1}(a)), which adds noise to model weights that are uninformative to the remaining data (training data excluding the concerning data) to achieve a weaker form of differential privacy \cite{dwork2014algorithmic}.

Different from previous work, we specifically consider the scenario of patient-wise forgetting, where instead of forgetting a selected random cohort of data, the data to be forgotten originate from a patient. We hypothesise (and show experimentally) that in a medical dataset, a patient's data can either be similar to other data (and form clusters) or form edge cases as we depict in Fig. \ref{fig:1}(b). These hypotheses are aligned with recent studies on long-tail learning \cite{buda2018systematic, liu2019large}, where different sub-populations within a class can exist with some being in the so-called long tail.\footnote{There is also a connection between edge cases and active learning \cite{settles2009active}, where one aims to actively label diverse data to bring more information to the model.} Subsequently, we will refer to these cases as \textit{common cluster} and \textit{edge case} hypotheses. 

We first study the patient-wise forgetting performance with simple translation of an existing machine unlearning method developed in \cite{golatkar2020eternal}. For patients under different hypotheses, forgetting and generalisation performance obtained after scrubbing \cite{golatkar2020eternal} vary as detailed in Section~\ref{sec:experiments}. In particular, the scrubbing method removes information not highly related to the remaining data to maintain good generalisation after forgetting, which is a weaker form of differential privacy \cite{dwork2014algorithmic}. When forgetting a patient under common cluster hypothesis, adequate performance can be achieved with the scrubbing method by carefully tuning the level of noise added to the model weights. When forgetting an edge-case patient, the scrubbing method does not remove specifically the edge-case patient's information but noise will be introduced to model weights corresponding to most of the edge cases in the remaining dataset. Hence, the overall model performance will be negatively affected. In fact, we observed in our experiment that data of a large portion of patients are edge cases while for computer vision datasets, the selected random cohort of data to be forgotten usually falls in the common cluster hypothesis. This limits the application of the scrubbing method and possibly other machine unlearning approaches that designed specifically for vision datasets to patient-wise forgetting.

To alleviate the limitation, we propose targeted forgetting, which only adds weighted noise to weights highly informative to a forgetting patient. In particular, we follow \cite{golatkar2020eternal} to measure the informativeness of model weights with Fisher Information Matrix (FIM), which determines the strength of noise to be added to different model weights. With the proposed targeted forgetting, we can precisely forget edge case data and maintain good model generalisation performance. For patient data fall under the common cluster hypothesis, the algorithm can forget their information with the trade-off of the model performance on the whole cluster. This implies that for some patients within the common cluster hypothesis, it is not easy to forget them without negatively affecting the model.

\noindent \textbf{Contributions}:
\begin{enumerate}
    \item We introduce the problem of patient-wise forgetting and formulate two hypotheses for patient-wise data.
    \item We show that machine unlearning methods specifically designed for vision datasets such as \cite{golatkar2020eternal} have poor performance in patient-wise forgetting. 
    \item We propose a new targeted forgetting method and perform extensive experiments on a medical benchmark dataset to showcase improved patient-wise forgetting performance.  
\end{enumerate}
Our work we hope will inspire future research to consider how different data affect forgetting methods especially in a patient-wise forgetting setting.

\section{Method}
Given a \textit{training} dataset $\mathcal{D}$, a \textit{forgetting} subset $\mathcal{D}_{f}\subset \mathcal{D}$ contains the images to be removed from a model $A(\mathcal{D})$, which is trained on $\mathcal{D}$ using any stochastic learning algorithm $A(\cdot)$. The \textit{retaining} dataset is the complement $\mathcal{D}_{r}$=$\mathcal{D}\setminus \mathcal{D}_{f}$, thus $\mathcal{D}_{r} \cap \mathcal{D}_{f}=\emptyset$. Test data is denoted as $\mathcal{D}_{test}$. For patient-wise forgetting, $\mathcal{D}_{f}$ is all the images of one patient. Let $\mathbf{w}$ be the weights of a model. Let $S(\mathbf{w})$ denote the operations applied to model weights to forget $\mathcal{D}_{f}$ in the model, and $A(\mathcal{D}_{r})$ be the \textit{golden standard} model. 

\subsection{The scrubbing method}
\label{sec:scubbing}
Assuming that $A(\mathcal{D})$ and $\mathcal{D}_{r}$ are accessible, Golatkar et al. \cite{golatkar2020eternal} propose a robust scrubbing procedure modifying model $A(\mathcal{D})$, to brings it closer to a golden standard model $A(\mathcal{D}_{r})$. They use FIM to approximate the hessian of the loss on $\mathcal{D}_{r}$, where higher values in FIM denote higher correlation between corresponding model weights and $\mathcal{D}_{r}$. With the FIM, they introduce different noise strength to model weights to remove information not highly informative to $\mathcal{D}_{r}$, and thus forget information corresponding to $\mathcal{D}_{f}$. The scrubbing function is defined as:
\begin{equation}
\mathrm{S}(\mathbf{w})=\mathbf{w}+\left(\lambda \sigma_{h}^{2}\right)^{\frac{1}{4}} F_{\mathcal{D}_{r}}(\mathbf{w})^{-1 / 4},
\end{equation}
where $F_{\mathcal{D}_{r}}(\mathbf{w})$ denotes the FIM computed for $\mathbf{w}$ on $\mathcal{D}_{r}$. Scrubbing is controlled by two hyperparameters: $\lambda$ decides the scale of noise introduced to $\mathbf{w}$ therefore it controls the model accuracy on $\mathcal{D}_{f}$; $\sigma_{h}$ is a normal distributed error term which simulates the error of the stochastic algorithm, ensuring a continuous gradient flow after the scrubbing procedure. Practically during experiments, the product of the two hyperparameters is tuned as a whole.

The Fisher Information Matrix $F$ of a distribution $P_{x, y}(\mathbf{w})$ w.r.t. $\mathbf{w}$ defined in \cite{golatkar2020eternal} is:
\begin{equation}
{F} = \mathbb{E}_{x \sim \mathcal{D}, y \sim p(y \mid x)}\left[\nabla_{\mathbf{w}} \log p_{\mathbf{w}}(y \mid x) \nabla_{\mathbf{w}} \log p_{\mathbf{w}}(y \mid x)^{T}\right]
\end{equation}

To save computational memory, only the diagonal values for FIM are computed and stored. The trace of FIM is calculated by taking the expectation of the outer product of the gradient of a deep learning model. In a medical dataset, the FIM ($F_{\mathcal{D}_{r}}(\mathbf{w})$) is derived by summing up the normalised FIM of each patient's data in the retaining set $\mathcal{D}_{r}$ and take the expectation at patient-level. Therefore, weights highly related to the cluster's features show high values in FIM because several cluster patients within $\mathcal{D}_{r}$ are correlated to these weights. Whereas for edge cases, no other patients are correlated with the same weights as of these edge cases; thus, the aggregated values in FIM for weights corresponding to edge cases are relatively small.

A value within $F_{\mathcal{D}_{r}}(\mathbf{w})$ reflects to what extent the change to its corresponding weights $\mathbf{w}$ would influence the model's classification process on this set $\mathcal{D}_{r}$. Hence, if a model weight is correlated with multiple data and thus considered to be important in classifying these data, its corresponding value in FIM would be relatively high, and vice versa. This also explains that weights correlated to data under common cluster hypothesis hold larger value than edge case hypothesis. Therefore, when scrubbing an edge case from a model, weights correlated to other edge cases even within $\mathcal{D}_{r}$ are also less informative to the remaining data thus will be scrubbed as well, making the model performance be negatively affected. 

\subsection{The targeted forgetting method}
\label{sec:targeted}
Based on the idea of scrubbing model weights, and the connection between the hessian of a loss on a set of data of a model and the extent to which the weights are informative about these data, we develop the targeted forgetting procedure. We assume access to the forgetting data $\mathcal{D}_{f}$ instead of $\mathcal{D}_{r}$. We believe that even in a real patient-wise forgetting scenario, temporary access to patient data is permissible until forgetting is achieved.

We compute FIM for $\mathbf{w}$ on $\mathcal{D}_{f}$ instead of $\mathcal{D}_{r}$ to approximate the noise added to model weights. Instead of keeping the most informative weights corresponding to $\mathcal{D}_{r}$ as in \cite{golatkar2020eternal}, our method precisely introduce noise to model weights highly informative about $\mathcal{D}_{f}$ (see Fig.~\ref{fig:1}(a)). Our proposed targeted forgetting is defined as:
\begin{equation}
\mathrm{S_T}(\mathbf{w})=\mathbf{w}+\left(\lambda_T \sigma_{h_T}^{2}\right)^{\frac{1}{4}} F_{\mathcal{D}_{f}}(\mathbf{w})^{1 / 4},
\end{equation}
where $\lambda_T$ and $\sigma_{h_T}$ are analogous parameters to $\lambda$ and $\sigma_h$ defined in Eq 1.

\noindent \textbf{Performance on the two hypotheses} \textit{Common cluster hypothesis}: Targeted forgetting will add noise to the most informative model weights corresponding to $\mathcal{D}_{f}$ so it will also reduce model performance on the corresponding cluster in $\mathcal{D}_{r}$ and $\mathcal{D}_{test}$. \textit{Edge case hypothesis}: Targeted forgetting will precisely remove information of an edge case and maintain good model performance. Results and discussion are detailed in Section \ref{sec:experiments}.

\section{Experiments}
\label{sec:experiments}
We first explore why scrubbing \cite{golatkar2020eternal} works well on computer vision datasets but shows poorer performance on patient-wise forgetting. We conduct an experiment to demonstrate the intrinsic dataset biases of CIFAR-10 \cite{krizhevsky2009learning} and ACDC \cite{bernard2018deep}. Then, we compare the forgetting and model performance after forgetting achieved using the scrubbing and our targeted forgetting methods. 

\begin{figure}[t]
\centering
\includegraphics[width=0.8\textwidth]{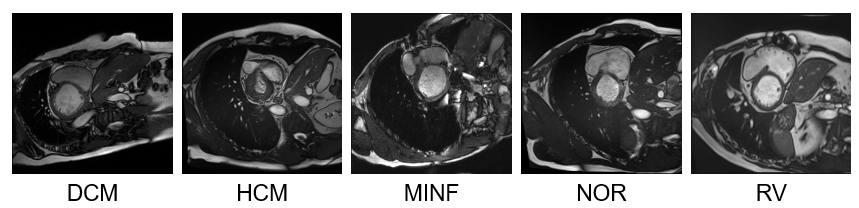}
\caption{Example images of ACDC dataset. DCM: dilated cardiomyopathy. HCM: hypertrophic cardiomyopathy. MINF: myocardial infarction. NOR: normal subjects. RV: abnormal right ventricle.} \label{fig:3:ACDC}
\end{figure}

\noindent \textbf{Datasets}: CIFAR-10 has 60,000 images (size $32\times 32$) of 10-class objects. The Automated Cardiac Diagnosis Challenge (ACDC) dataset contains 4D cardiac data from 100 patients with four pathologies classes and a normal group. We split the 100 patients into training and testing subsets. Overall, by preprocess the patient data into $224\times 224$ 2D images, there are 14,724 images from 90 patients form $D$, and 1,464 images from 10 patients form $\mathcal{D}_{test}$. Patients in both sets are equally distributed across the five classes. Example images from the ACDC dataset are shown in Fig.~\ref{fig:3:ACDC}. When conducting experiments under the patient-wise forgetting scenario, we only select one patient to be forgotten devising the forgetting set composed of all the images of the same patient.

\noindent \textbf{Implementation details}: For CIFAR-10, we follow the implementation steps in \cite{golatkar2020eternal}. When training the ACDC classifier, the model has a VGG-like architecture as in \cite{thermos2021miccai}. We use Cross Entropy as the loss function and use Adam optimizer \cite{kingma2014adam} with $\beta_1=0.5$, $\beta_2=0.999$. During training we use data augmentation including random rotation, Gaussian blur, horizontal and vertical flip. We train all classifiers with a learning rate of 0.0001 for 13 epochs. The original model trained with all 90 patients has $0.00$ error on $\mathcal{D}_{r}$ and  $\mathcal{D}_{f}$, and $0.19$ error on $\mathcal{D}_{test}$.

\begin{figure}[t]
\includegraphics[scale=0.38]{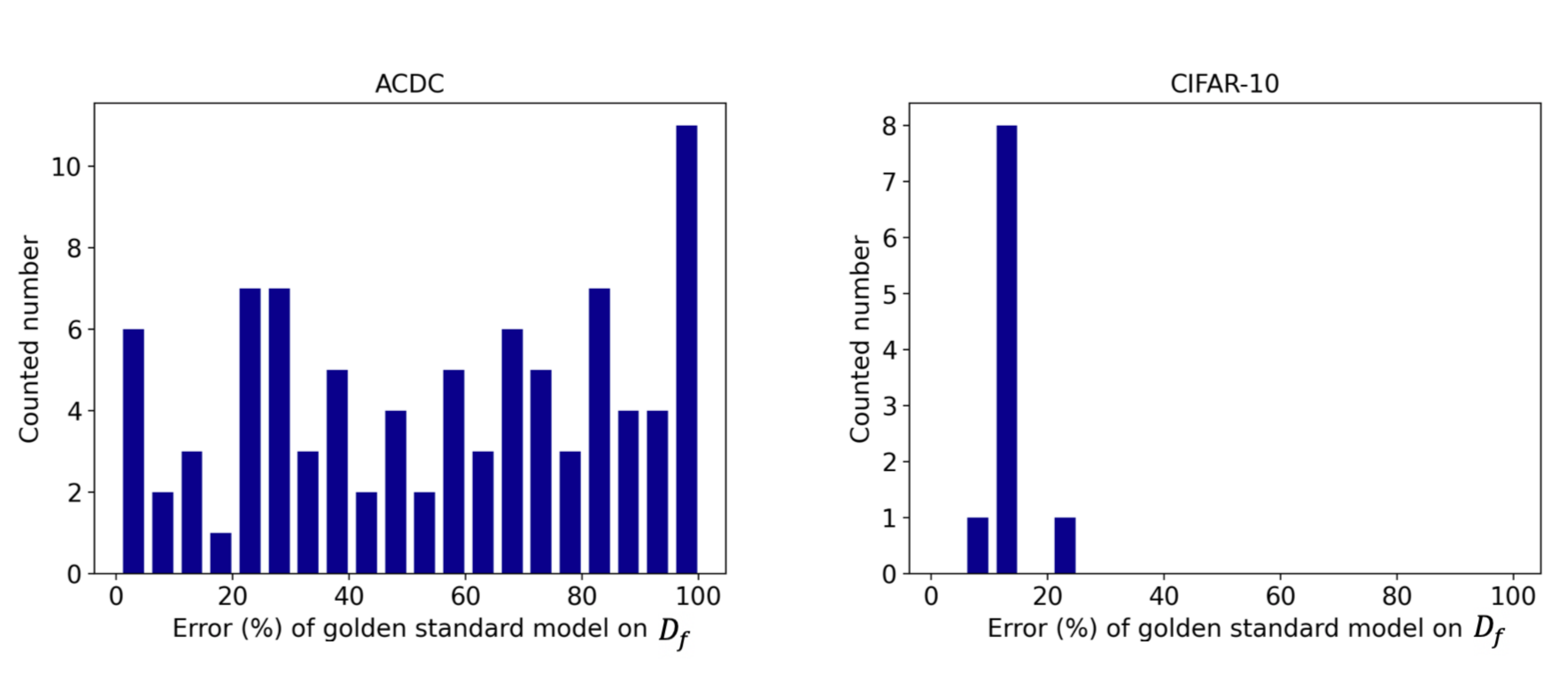}
\centering
 \caption{Histograms of re-training experiments. The y-axis refers to the total number of patients/sets whose golden standard lies within an interval(e.g. [95,100]) of x-axis.}
 \label{1}
\end{figure}
\subsection{The hardness of patient-wise forgetting}
\label{sec:hardness}
Here we compare between CIFAR-10 and ACDC to show that some patient data are hard to learn and forget. For 90 patients in ACDC, we remove one patient's data and re-train a model on the remaining 89 to be the golden standard model, $A(\mathcal{D}_{r})$. We then measure the error of the deleted patient on $A(\mathcal{D}_{r})$. We repeat this for all 90 patients. For CIFAR-10, we select 10 non-overlapping sets from its training set, each with 100 images from the same class, to be the deleting candidates and repeat the re-train experiments. Data are hard to generalize by its golden model show high error on $A(\mathcal{D}_{r})$ and thus should be hard to forget.

\noindent \textbf{Results and discussion}: 
Fig.~\ref{1} collects the findings of this experiment as histograms and shows the differences between the two datasets. Overall, for ACDC, the 90 individually measured results of classification error of a $\mathcal{D}_{f}$ on its corresponding golden model $A(\mathcal{D}_{r})$ vary from 0$\%$ to 100$\%$, whereas in CIFAR-10, the 10 experimental results only vary from 10$\%$ to 25$\%$. High golden model error on a $\mathcal{D}_{f}$ means that the model is unable to generalise to this patient's data; thus, this patient is not similar to any other patients in the training set, and must belong to the edge case hypothesis. By considering a threshold of $50\%$ on the error of the golden model, we find that \textbf{$>60\%$ of patients in ACDC can be considered to belong to the edge case hypothesis}. This is remarkably different in CIFAR-10: golden model results concentrate at low error indicating that few edge cases exist. In addition, as discussed in section \ref{sec:scubbing}, when dealing with edge cases, scrubbing can degrade model performance. This will explain the results of the scrubbing method: under-performance in ACDC because many patients fall under the edge case hypothesis. 

\begin{table}[t]
\centering
\caption{Forgetting results for four patients. We report Error $=1-$Accuracy on the forgetting ($\mathcal{D}_{f}$) and test ($\mathcal{D}_{test}$) sets respectively. Scrubbing Method refers to the method of \cite{golatkar2020eternal} whereas Targeted Forgetting refers to the method in Section \ref{sec:targeted}. \color{red}Red\color{black} \ and \color{blue}blue\color{black} \ denote the golden standard of forgetting performance for each row respectively, with performance being better when it is closer to the standard. With respect to error on $\mathcal{D}_{f}$ \textbf{High} noise level refers to the noise strength when a method reaches 1.00 error;  \textbf{Medium}: 0.85$\pm$0.05 error; and \textbf{Low}: 0.14$\pm$0.05 error. The confidence bar is obtained over three experiments.}
\label{table2}
\resizebox{\textwidth}{!}{%
\begin{tabular}{|c|c|c|cccccc|}
\hline
\multirow{3}{*}{\begin{tabular}[c]{@{}c@{}}Patient \\ ID\end{tabular}} & \multirow{3}{*}{Error on}                                                    & \multirow{3}{*}{\begin{tabular}[c]{@{}c@{}}Golden \\ Standard\end{tabular}} & \multicolumn{6}{c|}{Noise level} \\ \cline{4-9} 
&&& \multicolumn{2}{c|}{Low}& \multicolumn{2}{c|}{Medium}& \multicolumn{2}{c|}{High}\\ \cline{4-9} 
&&& \multicolumn{1}{c|}{Scrubbing}& \multicolumn{1}{c|}{\begin{tabular}[c]{@{}c@{}}Targeted \\ Forgetting\end{tabular}} 
& \multicolumn{1}{c|}{Scrubbing}& \multicolumn{1}{c|}{\begin{tabular}[c]{@{}c@{}}Targeted \\ Forgetting\end{tabular}} 
& \multicolumn{1}{c|}{Scrubbing}& \begin{tabular}[c]{@{}c@{}}Targeted \\ Forgetting\end{tabular} \\ \hline
\begin{tabular}[c]{@{}c@{}}94\\ (Edge)\end{tabular} & \begin{tabular}[c]{@{}c@{}}$\mathcal{D}_{f}$\\ $\mathcal{D}_{test}$\end{tabular} 
& \begin{tabular}[c]{@{}c@{}}\color{blue}1.000$\pm$0.000\\ \color{red}0.237$\pm$0.002\end{tabular}                         
& \multicolumn{1}{c|}{\begin{tabular}[c]{@{}c@{}}0.154$\pm$0.005\\ 0.671$\pm$0.012\end{tabular}} 
& \multicolumn{1}{c|}{\begin{tabular}[c]{@{}c@{}}0.174$\pm$0.020\\ 0.223$\pm$0.011\end{tabular}}            
& \multicolumn{1}{c|}{\begin{tabular}[c]{@{}c@{}}0.859$\pm$0.010\\ 0.739$\pm$0.007\end{tabular}} 
& \multicolumn{1}{c|}{\begin{tabular}[c]{@{}c@{}}0.851$\pm$0.018\\ 0.291$\pm$0.005\end{tabular}}            
& \multicolumn{1}{c|}{\begin{tabular}[c]{@{}c@{}}1.000$\pm$0.000\\ 0.746$\pm$0.008\end{tabular}} 
& \begin{tabular}[c]{@{}c@{}}1.000$\pm$0.000\\ 0.316$\pm$0.002\end{tabular}            \\ \hline
\begin{tabular}[c]{@{}c@{}}5\\ (Edge)\end{tabular}& \begin{tabular}[c]{@{}c@{}}$\mathcal{D}_{f}$\\ $\mathcal{D}_{test}$\end{tabular} 
& \begin{tabular}[c]{@{}c@{}}\color{blue}0.809$\pm$0.009\\ \color{red}0.253$\pm$0.026\end{tabular}                         
& \multicolumn{1}{c|}{\begin{tabular}[c]{@{}c@{}}0.127$\pm$0.022\\ 0.394$\pm$0.017\end{tabular}} 
& \multicolumn{1}{c|}{\begin{tabular}[c]{@{}c@{}}0.121$\pm$0.019\\ 0.269$\pm$0.004\end{tabular}}            
& \multicolumn{1}{c|}{\begin{tabular}[c]{@{}c@{}}0.853$\pm$0.020\\ 0.624$\pm$0.015\end{tabular}} 
& \multicolumn{1}{c|}{\begin{tabular}[c]{@{}c@{}}0.857$\pm$0.002\\ 0.407$\pm$0.001\end{tabular}}            
& \multicolumn{1}{c|}{\begin{tabular}[c]{@{}c@{}}0.997$\pm$0.003\\ 0.696$\pm$0.002\end{tabular}} 
& \begin{tabular}[c]{@{}c@{}}1.000$\pm$0.000\\ 0.506$\pm$0.002\end{tabular}            \\ \hline
\begin{tabular}[c]{@{}c@{}}13\\ (Cluster)\end{tabular} & \begin{tabular}[c]{@{}c@{}}$\mathcal{D}_{f}$\\ $\mathcal{D}_{test}$\end{tabular} 
& \begin{tabular}[c]{@{}c@{}}\color{blue}0.202$\pm$0.004\\ \color{red}0.194$\pm$0.012\end{tabular}                         
& \multicolumn{1}{c|}{\begin{tabular}[c]{@{}c@{}}0.111$\pm$0.006\\ 0.361$\pm$0.001\end{tabular}} 
& \multicolumn{1}{c|}{\begin{tabular}[c]{@{}c@{}}0.092$\pm$0.002\\ 0.343$\pm$0.007\end{tabular}}            
& \multicolumn{1}{c|}{\begin{tabular}[c]{@{}c@{}}0.871$\pm$0.018\\ 0.590$\pm$0.005\end{tabular}} 
& \multicolumn{1}{c|}{\begin{tabular}[c]{@{}c@{}}0.850$\pm$0.021\\ 0.524$\pm$0.013\end{tabular}}            
& \multicolumn{1}{c|}{\begin{tabular}[c]{@{}c@{}}1.000$\pm$0.000\\ 0.694$\pm$0.004\end{tabular}} 
& \begin{tabular}[c]{@{}c@{}}1.000$\pm$0.000\\ 0.602$\pm$0.016\end{tabular}            \\ \hline
\begin{tabular}[c]{@{}c@{}}9\\ (Cluster)\end{tabular} & \begin{tabular}[c]{@{}c@{}}$\mathcal{D}_{f}$\\ $\mathcal{D}_{test}$\end{tabular} 
& \begin{tabular}[c]{@{}c@{}}\color{blue}0.010$\pm$0.002\\ \color{red}0.233$\pm$0.007\end{tabular}                         
& \multicolumn{1}{c|}{\begin{tabular}[c]{@{}c@{}}0.176$\pm$0.005\\ 0.402$\pm$0.012\end{tabular}} 
& \multicolumn{1}{c|}{\begin{tabular}[c]{@{}c@{}}0.152$\pm$0.009\\ 0.442$\pm$0.001\end{tabular}}            
& \multicolumn{1}{c|}{\begin{tabular}[c]{@{}c@{}}0.892$\pm$0.003\\ 0.643$\pm$0.006\end{tabular}} 
& \multicolumn{1}{c|}{\begin{tabular}[c]{@{}c@{}}0.862$\pm$0.005\\ 0.613$\pm$0.001\end{tabular}}            
& \multicolumn{1}{c|}{\begin{tabular}[c]{@{}c@{}}0.998$\pm$0.002\\ 0.699$\pm$0.005\end{tabular}} 
& \begin{tabular}[c]{@{}c@{}}0.995$\pm$0.005\\ 0.656$\pm$0.001\end{tabular}            \\ \hline
\end{tabular}%
}
\end{table}

\subsection{Patient-wise forgetting performance}
\label{sec:without}
We focus on four representative patients using the analysis in Section \ref{sec:hardness}: patients 94 and 5 that fall under the edge case hypothesis; and patients 13 and 9 fall under a common cluster hypothesis. Here we consider a stringent scenario: the re-trained golden standard model is not available for deciding how much to forget, so the level of noise to be added during scrubbing or forgetting is unknown. We adjust noise strength (low, medium and high) by modulating the hyperparameters in both methods to achieve different levels of forgetting.\footnote{For our experiments we fix to introduce noise to 1$\%$ most informative weights (based on extensive experiments) when applying the targeted forgetting.} We assess forgetting performance by comparing against golden standard models: A method has good forgetting performance by coming as close to the performance of the golden standard model on $\mathcal{D}_{test}$.

\begin{table}[htbp]
\centering
\caption{The average noise value added to weights at High (when achieving 1.00 error on $\mathcal{D}_{f}$). Note that medium and low noise is with 66.7$\%$ and  30.0$\%$ of high noise level respectively. }
\label{tab:table5}
\resizebox{0.5\textwidth}{!}{%
\begin{tabular}{|c|cc|}
\hline
\multirow{2}{*}{\begin{tabular}[c]{@{}c@{}}Patient \\ ID\end{tabular}} & \multicolumn{2}{c|}{High}                            \\ \cline{2-3} 
                                                                       & \multicolumn{1}{c|}{Scrubbing} & Targeted Forgetting \\ \hline
94 (Edge)                                                              & \multicolumn{1}{c|}{2.33E-05}  & 3.00E-06            \\ \hline
5 (Edge)                                                               & \multicolumn{1}{c|}{1.65E-05}  & 4.5E-06             \\ \hline
13 (Cluster)                                                           & \multicolumn{1}{c|}{1.6E-05}   & 8.66E-06            \\ \hline
9 (Cluster)                                                            & \multicolumn{1}{c|}{1.43E-05}  & 1.2E-05             \\ \hline
\end{tabular}%
}
\end{table}

\noindent \textbf{Is targeted forgetting better for forgetting edge cases?}
For edge cases, forgetting can be achieved (compared to the golden standard) at high level of noise with both methods. However, the scrubbing method significantly degrades the model generalisation performance. With targeted forgetting, good model generalisation performance on $\mathcal{D}_{test}$ at all noise levels is rather maintained. Additionally from Table \ref{tab:table5} we observe that the scrubbing method adds more noise to model weights to forget an edge case. This further supports our discussion in section \ref{sec:scubbing} on how the scrubbing method negatively affects the overall model performance when forgetting edge case.

\noindent \textbf{Is targeted forgetting better for forgetting common cluster cases?}
For common cluster cases, both methods can achieve standard forgetting with a near low level of noise with nice model's generalisation performance on $\mathcal{D}_{test}$, as shown in Table \ref{table2}.
For example for patient 13, the test error of two methods at low noise level is $0.361$ and $0.343$, which is close and relatively small. When the noise level grows to medium and high to forget more, although the test error with two methods still being close, it grows to a high value. Overall, when forgetting common cluster cases, the two methods show similar good performance at a standard level of forgetting and they both can forget more about a patient by sacrificing the model's generalisation.

\noindent \textbf{Can patient data be completely forgotten?} Overall, for edge cases, using targeted forgetting, the patient-wise data can be completely forgotten (achieving error higher than 0.80 (random decision for 5 classes in our case) on $\mathcal{D}_{f}$) without sacrificing the model generalisation performance. While for common cluster cases, it is less likely to forget the patient data as completely forgetting will result the significantly degraded generalisation performance with the scrubbing or our targeted forgetting. In fact, the level of noise added to the model weights affects the trade-off between model performance and respecting data protection. Higher noise leads to more information being removed, thus protecting the data better yet degrading the model accuracy. Therefore, the noise needs to be carefully designed such that a sweet spot between forgetting and generalisation performance can be achieved.

\section{Conclusion}
We consider patient-wise forgetting in deep learning models. Our experiments reveal that forgetting a patient's medical image data is harder than other vision domains. We found that this is due to data falling on two hypotheses: common cluster and edge case. We identified limitations of an existing state-of-the-art scrubbing method and proposed a new targeted forgetting approach. Experiments highlight the different roles of these two hypotheses and the importance of considering the dataset bias. We perform experiments on cardiac MRI data but our approach is data-agnostic, which we plan to apply on different medical datasets in the future. In addition, future research on patient-wise forgetting should focus on better ways of detecting which hypothesis the data of patients belong to and how to measure patient-wise forgetting performance with considering the two hypotheses. 

\section{Acknowledgements} 
This work was supported by the University of Edinburgh, the Royal Academy of Engineering and Canon Medical Research Europe by a PhD studentship to Xiao Liu. This work was partially supported by the Alan Turing Institute under EPSRC grant EP/N510129/1. S.A. Tsaftaris acknowledges the support of Canon Medical and the Royal Academy of Engineering and the Research Chairs and Senior Research Fellowships scheme (grant RC-SRF1819$\backslash$8$\backslash$25) and the [in part] support of the Industrial Centre for AI Research in digitalDiagnostics  (iCAIRD, https://icaird.com)  which  is  funded  by  Innovate  UK  on  behalf  of  UK  Research  and Innovation (UKRI) [project number:  104690].

\bibliographystyle{splncs04}
\bibliography{bibliography}

\section{Appendix}

\begin{figure}[ht]
\centering
\setcounter{figure}{0} 
\renewcommand{\thefigure}{A\arabic{figure}}
\includegraphics[width=0.6\textwidth]{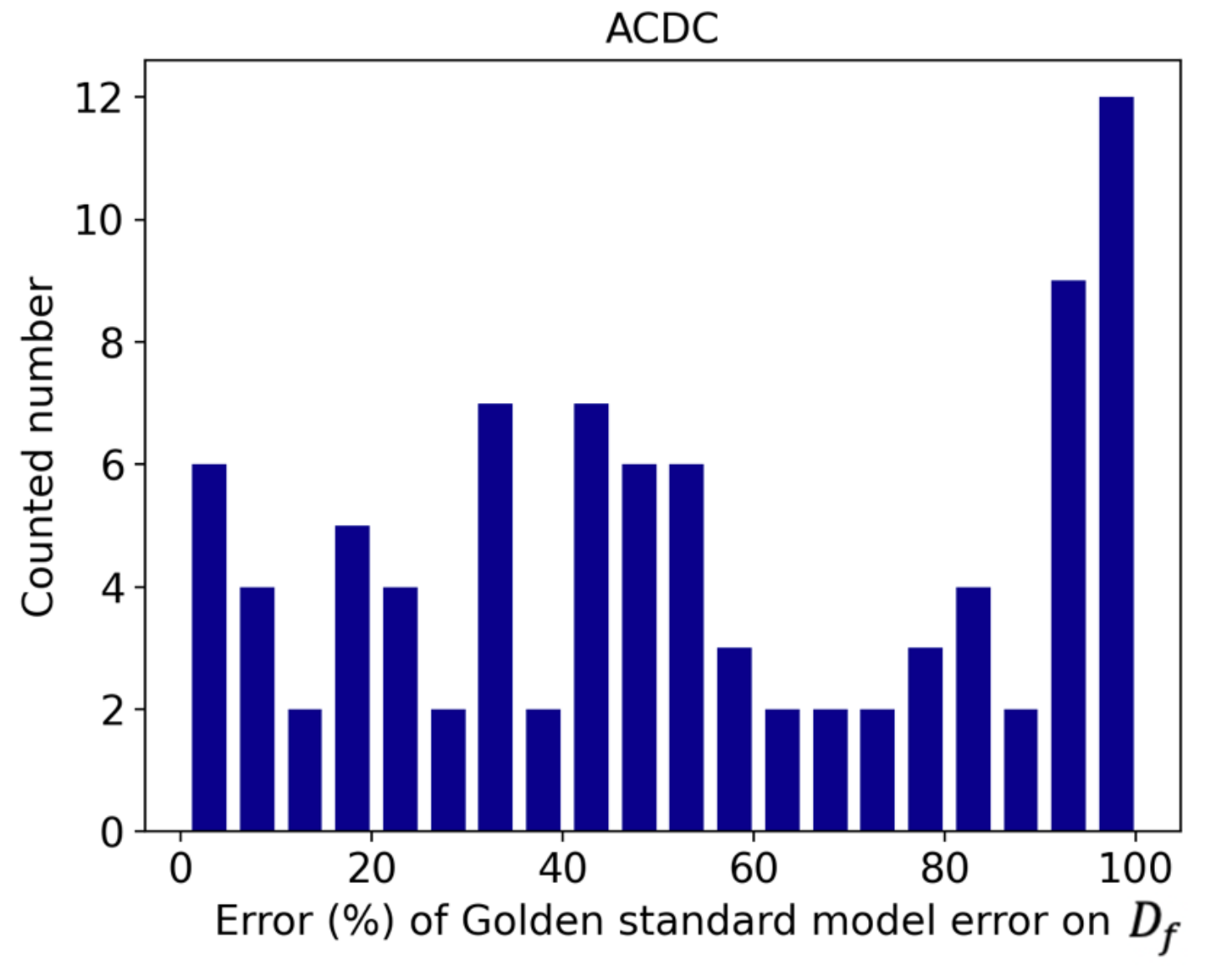}
\caption{Histograms of re-training experiments. The y-axis refers to the total number of patients/sets whose golden standard lies within an interval(e.g. [95,100]) of x-axis.} 
\label{fig:appD}
\end{figure}

We explore if overfitting would be an issue affecting the results in Section 3.1 by redoing the experiment by early-stop training models. With all the settings being the same as in Section 3, the training epochs for the 90 individual models is changed from 13 to 7 to obtain less overfitted models. 

Fig. \ref{fig:appD} collects the results with early stop models. Overall, compared with Fig.3 in Section 3.1, although the distribution of the early stop results histogram is slightly different, the 90 individually measured results of classification error of a $\mathcal{D}_{f}$ on its corresponding golden model $A(\mathcal{D}_{r})$ also vary from 0$\%$ to 100$\%$. By considering a threshold of $50\%$ on the error of the golden model, there are still \textbf{$>50\%$} of patients in ACDC can be considered to belong to the edge case hypothesis. Therefore, overfitting is not considered the reason for the emergence of edge cases.

\end{document}